\documentclass{article}
\usepackage{spconf,amsmath,graphicx}
\usepackage{comment} 
\usepackage{algorithm}
\usepackage{booktabs}
\usepackage{multirow}
\usepackage{subcaption}
\usepackage{multirow}
\usepackage[noend]{algpseudocode}
\usepackage{color, colortbl}
\usepackage{float}
\usepackage[utf8]{inputenc}
\usepackage{dblfloatfix} 

\usepackage{amssymb}

\makeatletter
\algnewcommand{\LineComment}[1]{\Statex \hskip\ALG@thistlm \(\triangleright\) #1}
\makeatother

\title{Continual learning of predictive models in video sequences via variational autoencoders}
\name{Damian Campo, Giulia Slavic, Mohamad Baydoun, Lucio Marcenaro, Carlo Regazzoni\vspace{-0.15cm}}

\address{\vspace{-0.15cm}DITEN, University of Genova, Italy}
\begin{document}
\ninept
\maketitle
\begin{abstract}
This paper proposes a method for performing continual learning of predictive models that facilitate the inference of future frames in video sequences. For a first given experience, an initial Variational Autoencoder, together with a set of fully connected neural networks are utilized to respectively learn the appearance of video frames and their dynamics at the latent space level. By employing an adapted Markov Jump Particle Filter, the proposed method recognizes new situations and integrates them as predictive models avoiding catastrophic forgetting of previously learned tasks. For evaluating the proposed method, this article uses video sequences from a vehicle that performs different tasks in a controlled environment.  
\end{abstract}
\vspace{-0.25cm}
\section{Introduction} \label{sec:intro}
\vspace{-0.25cm}
Some biological organisms, such as pigeons and large primates, possess the ability to learn new experiences continuously through their lifetime \cite{Fagot2006, Shanahan2013, Milne2018}. Such a capacity of preserving and use information from past experiences allows organisms to develop cognitive skills that are often critical for survival \cite{Beran2016, Darby2018}. As discussed in \cite{Dukas2019}, the development of biological expertise follows a characteristic pattern of gradual improvement of performance over a particular task. The work in \cite{Dukas2019} also claims that the level of expertise reached by an organism depends on its \textit{i)} long-term memory, \textit{ii)} working memory capacity, \textit{iii)} ability to focus attention on relevant information, \textit{iv)} capability to anticipate, perceive and comprehend surroundings, \textit{v)} velocity at the decision-making and \textit{vi)} coordination in motor movements. The capabilities mentioned above can be seen as a set of cognitive abilities that organisms employ to solve problems and adapt to new situations systematically. We argue that each of those skills can be improved/refined by recalling past experiences that match with current situations, suggesting a major role of continual learning when solving problems and developing expertise.                

Motivated by the aforementioned research studies on the continual learning in living beings, we propose a method by which video sequences acquired by artificial systems are employed to create models that can predict future situations based on past experiences. The proposed method facilitates the continual learning of new experiences by using abnormal information (video-frames) detected from available predictive models. We believe artificial systems can be highly benefited by the continual learning of new experiences, contributing to the automatic development of cognitive skills that facilitate the growth of expertise and adaptability in machines.  

Several articles have tried to include continual learning (also known as lifelong learning) capabilities into artificial systems inspired by findings in psychology and neuroscience \cite{Zenke2017, Flesch2018}. Primarily, continual learning has been studied in deep neural networks (DNNs) due to their remarkable advances across diverse applications \cite{Parisi2019}. Nonetheless, when trying to integrate new information to artificial neural networks (ANNs), it is observed a dramatic performance degradation in previously learned tasks, such a phenomenon is known as \textit{catastrophic forgetting}; and various articles have tried to overcome it by using different techniques \cite{Li2017, Kirkpatrick2017, Yao2019}. In particular, the work in \cite{Zenke2017} distinguishes three main approaches to deal with catastrophic forgetting, facilitating the continual learning in ANNs: \textit{i) Architectural}, where the architecture of the network is modified, and the objective function remains the same. \textit{ii) Functional}, which modifies the loss function, encouraging that the learning of new tasks does not affect already learned information. \textit{iii) Structural}, consisting of introducing penalties on parameters to avoid forgetting already learned experiences. 

The proposed method consists of an architectural approach that enables the continual learning of predictive models in video sequences by adopting a duplication and tuning process over Variational Autoencoders (VAEs), which are trained as new situations are detected. For a given task, our work uses a VAE for learning the appearance features of video data and a set of fully connected ANNs for performing predictions of following video frames at the latent space level. Our work demonstrates how predictive models can be learned and used incrementally as new situations are observed.  

Although various research works have tackled the problem of continual learning in DNNs, their contributions have been mainly focused on classifications tasks \cite{Rebuffi2017, Nguyen2017, Lesort2019} consisting in continuously learning different classes from diverse datasets such as CIFAR-10/100, ImageNet, and MNIST. On the other hand, our work focuses on the ability to predict the subsequent frames of video sequences by associating visual observations with already identified/learned experiences. Moreover, the proposed method employs a Markov Jump Particle Filter (MJPF) \cite{Baydoun_fusion} over latent space information for making predictions that can be potentially integrated with other sensory information, e.g., positional and control data.       

The main contributions of the proposed method are: \textit{i)} The employment of VAEs that facilitate to obtain latent spaces from which to create predictive models at the low-dimensional level. \textit{ii)} The detection of new situations that autonomous systems may employ to learn continuously predictive models without forgetting previously learned experiences.  \textit{iii)} For evaluation purposes, this paper uses real video sequences coming from a vehicle performing different tasks in a controlled environment.

The rest of the paper is organized as follows: Section \ref{sec:method} explains the proposed method for enabling continual learning over video sequences based on VAEs' latent spaces. Section \ref{sec:dataset} introduces the employed dataset. Section \ref{sec:exp_results} discusses the obtained results and section \ref{sec:conclusion} concludes the article and suggests future developments. 

\begin{figure*}[ht]
\begin{center}
\includegraphics[width=\textwidth]{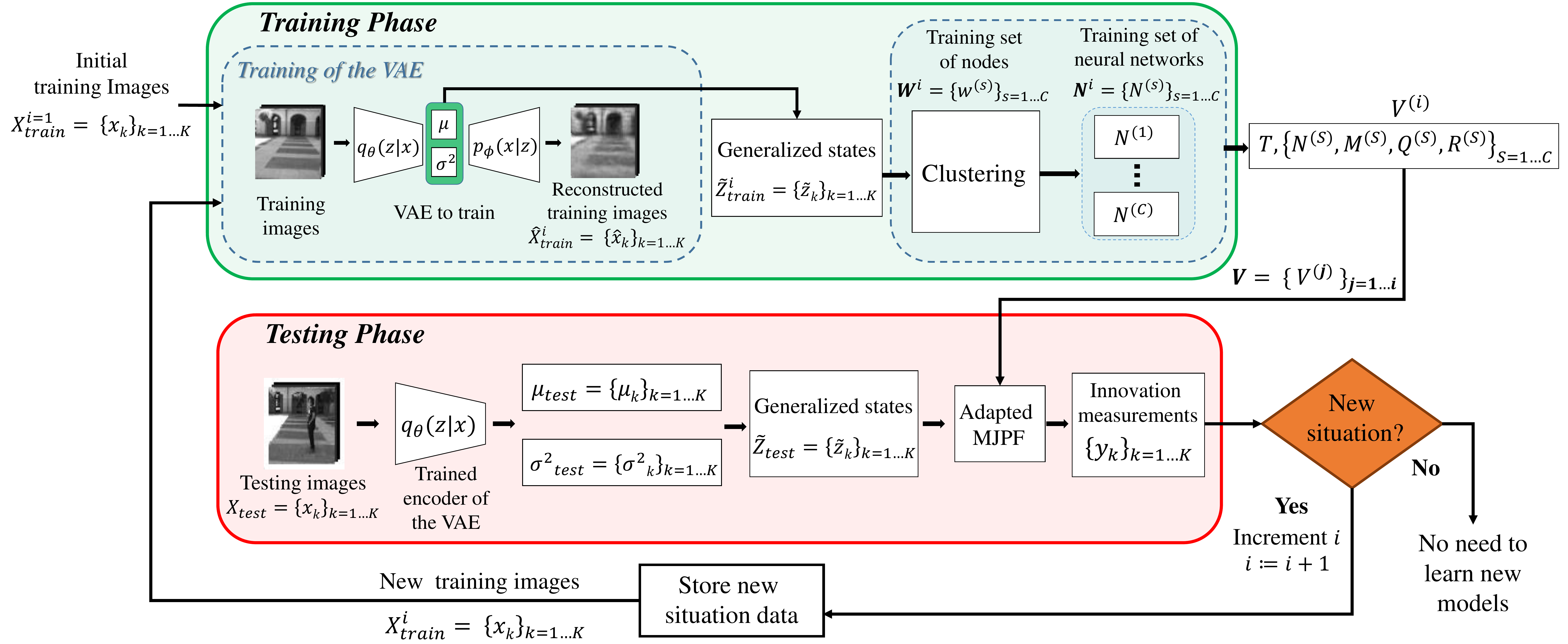}
\vspace{-0.6cm}
\caption{Block diagram of proposed method.}
\vspace{-0.8cm}
\label{fig:BlockDiagram}
\end{center}
\end{figure*}

\vspace{-0.25cm}
\section{Method}\label{sec:method}
\vspace{-0.25cm}
The proposed method includes two major phases that are continuously repeated each time new situations are detected: \emph{i)} a training process (section~\ref{trainingPhase}) to learn probabilistic models based on observed data, and an online testing procedure (section~\ref{testingPhase}) for detecting possible new situations on observed data. Accordingly, when new situations are identified, they are employed to trigger a new training process that refines available predictive models (section~\ref{continual_learning}). The proposed method is summarized in the block diagram shown in Fig.~\ref{fig:BlockDiagram}.



\vspace{-0.35cm}
\subsection{Training phase}\label{trainingPhase}
\vspace{-0.1cm}
{\bf Variational Autoencoder}. A Variational Autoencoder (VAE) is used for describing images in a latent space that significantly reduces the original dimension of video frames. Moreover, a VAE facilitates to represent images in the latent state probabilistically by using a mean $\mu$ and variance $\sigma^2$ to approximate each latent variable. 
As is well known, a VAE is composed of two parts: an encoder $q_{\theta}(z|x)$ and a decoder $p_{\phi}(x|z)$. The latent state $z$ sampled from $\mathcal{N}( \mu,\sigma^{2} )$, returns an approximate reconstruction of the observation $x$. Through $\theta$ and $\phi$, we define the parameters of both encoder and decoder, respectively. To optimize them, the VAE maximizes the sum of the lower bound on the marginal likelihood of each observation $x$ of the dataset $D$, as described in \cite{kingma2014autoencoding,Kingma2019}.





This work uses the VAE's ability to encode visual information into a significant lower-dimensional probabilistic latent space, which is employed to make inferences of future instances. Consequently, we first train the VAE with a set of training images $X^i_{train}$, where $i$ indexes the task to be learned (initially, $i = 1$). By utilizing the trained VAE's encoder, we obtain a set of latent features described by $\mu_{train}$ and $\sigma^{2}_{train}$, which represent $X^i_{train}$ data.\\
{\bf Generalized states}. Let $\mu_{train}$ be a set training images' states corresponding to the VAE's latent space data; we build a set of Generalized States (GSs) containing also the first-order time derivatives of $\mu_{train}$. Accordingly, let $\mu_{k} \in \mu_{train}$ be the state of the training image at time $k$, its first-order time derivative can be approximated by $\dot{\mu}_{k} \sim \frac{\mu_{k} - \mu_{k-1}}{\Delta k}$, $\dot{\mu}_{k} \in \dot{\mu}_{train}$. $\Delta k = 1$ assumes a normalized regular sampling of images. The GS at time $k$ can thus be written as $\tilde{z}_{k} =[\mu_{k} \hspace{0.2cm} \dot{\mu}_{k}]^\intercal$. By repeating this for each image $x_k \in X^i_{train}$, we obtain a set of GSs for the training set, defined by:
\begin{equation}\label{eq:GSs_train}
\tilde{Z}_{train}^i = [\mu_{train} \hspace{0.2cm} \dot{\mu}_{train}]^\intercal.
\end{equation}
{\bf Clustering and neural networks}. After obtaining $\tilde{Z}_{train}^i$, we use a traditional k-means algorithm to cluster GSs into groups that carry similar information. Since we use $\mu$ and $\dot{\mu}$ as input data, obtained clusters capture information of encoded images and their dynamics. 

By letting $C$ be the total number of identified clusters, it is possible to use $S$ to index clusters, such that $S \in \{1,\dots,C\}$. Once the clustering is performed, we calculate a transition matrix $T$ encoding the passage probabilities from each cluster to the others. Consequently, the following features are extracted from each cluster $S$: \textit{i)} cluster's centroid $M^{(S)}$, \textit{ii)} cluster's  covariance $Q^{(S)}$ and \textit{iii)} cluster's radius of acceptance $R^{(S)}$. Finally, a fully connected neural network $N^{(S)}$ defining the dynamics of GSs, i.e., continuous predictive model, is learned for each cluster. For training each $N^{(S)}$, the value of every $\mu_{k}$ is taken as input and the corresponding $\dot{\mu}_{k+1}$ as output, where $[\mu_k, \dot{\mu}_{k}]^\intercal \in S$, such that:
\begin{equation}\label{eq:NN_eq}
\dot{\mu}_{k+1} \sim N^{(S)}(\mu_{k}) + w_{k},
\end{equation}
where $w_{k}$ is the residual error after the convergence of the network.





Each $N^{(S)}$ learns a sort of \textit{quasi-semantic} information based on a particular image appearance and motion detected by the cluster $S$, facilitating the estimation of future latent spaces, i.e., predicting the following frames. Such predictions can be employed to measure the similarity between new observations and previously learned experiences encoded into NNs. In case predictions from NNs are not compliant with observations, an abnormality should be detected, and models should be adapted to learn new situations and semantic information. Consequently, each identified task $i$ can be described by the set of parameters $V^{(i)} = \Big\{ T, \{N^{(S)}, M^{(S)}, Q^{(S)}, R^{(S)}\}_{S=1,\dots,C} \Big\}$.

\vspace{-0.35cm}
\subsection{Testing Phase}\label{testingPhase}
\vspace{-0.1cm}
During the testing phase, each image $X_{test}$ is processed through the VAE, and their respective GSs are calculated. Then, an adapted version of the MJPF based on the learned information $V^{(i=1)}$ is used to detect new situations in video sequences.\\
{\bf Adapted Markov Jump Particle Filter}. An MJPF, firstly proposed in \cite{Baydoun_fusion}, is adapted for prediction and abnormality detection purposes on visual data. The MJPF uses a Particle Filter coupled with a bank of Kalman Filters (KFs) for inferring continuous and discrete level information. Since this work tackles a problem that requires a non-linear predictive model and a non-linear observation model, solved respectively by the set of NNs and a VAE, it is employed a bank of unscented KFs (UKF) and VAE's encoded information for making inferences over video sequences. 


The proposed adapted MJPF (A-MJPF) follows two stages at each time instant $k$: \textit{prediction} and \textit{update}. During prediction, the next cluster $S_{k+1}$ (discrete level) and GS $\tilde{z}_{k+1}$ (continuous level) are estimated for each particle, i.e., $p(S_{k+1} | S_{k})$ and $p(\tilde{z}_{k+1}| \tilde{z}_{k})$ respectively. The prediction at discrete level is similar to the standard MPJF in \cite{Baydoun_fusion}. Instead, the A-MPJF uses the neural network $N^{(S_k)}$ to make predictions at a continuous level. Since non-linear models are considered for predicting continuous level information, a UKF is utilized as described in \cite{Wan} by taking $2L$ additional sigma points. Each sigma point's prediction follows the equation below: 
\begin{equation}\label{eq:prediction}
\tilde{z}\mathrm{_{k+1}} = f(\tilde{z}\mathrm{_{k}}) = A\tilde{z}\mathrm{_{k}} + BN\mathrm{^{(S)}}(\mu\mathrm{_{k}}) + w\mathrm{_k},
\end{equation}
$A$ and $B$ are matrices that map the previous state $\tilde{z}^i_k$ and the predicted velocity computed by $N^{(S)}(\mu^{i}_{k})$ on the new state $\tilde{z}^i_{k+1}$. $A = [A_1 A_2]$ with $A_1 = [I_L 0_{L,L}]^\intercal$, $A_2 = 0_{2L, L}$; and $B = [I_L I_L]^\intercal$. The mean and covariance of $\tilde{z}^i_{k+1}$ are calculated through the UKF.

The update phase is performed when a new measurement (image) is observed. At the discrete level, particles are resampled based on an innovation measurement. At the continuous state level, a modified KF is in charge of the update. This update takes into consideration the fact that $\mu_k$ and $\sigma_k^2$ associated with $x_k$ can be used as the mapped observation on the state space at time $k$. Consistently, $\sigma_k^2$ can approximate the covariance matrix, such that $\Sigma_{k} \sim I_{L}\sigma^{2}_k$, representing the uncertainty while encoding images. By assuming a negligible observation noise, it is possible to employ a modified version of the KF update equations where the observation matrix $H$ disappears. Algorithm \ref{algorithm_KF} describes the employed KF's steps.
\vspace{-0.2cm}
\begin{algorithm} 
\caption{Equations for the prediction and update phases of the Adapted Kalman Filter.}
\begin{algorithmic}[1]
\LineComment \textbf{PREDICTION}: 
\State Calculation of the sigma points $\tilde{z}^i_{k|k}$ and of their respective weights $\tilde{W}^{i, m}$ and $\tilde{W}^{i, c}$ as described in \cite{Wan}. 
\State $\tilde{z}^i_{k+1|k} = f(\tilde{z}^i_{k|k})$ 
\State $\tilde{z}_{k+1|k} = \sum_{i = 0}^{2L}\tilde{W}^{i, m}\tilde{z}^i_{k+1|k}$ 
\State $P_{k+1|k} = \sum_{i=0}^{2L}\tilde{W}^{i,c}\{\tilde{z}^i_{k+1|k} - \tilde{z}_{k+1|k}\}\{\tilde{z}^i_{k+1|k} - \tilde{z}_{k+1|k}\}^\intercal$
\State $P^L_{k+1|k} = P_{k+1|k}\Big|_{\{row: 1...L, col: 1...L\}}$
\vspace{0.15cm}
\LineComment \textbf{UPDATE}: 
\State $K_{k+1} = [P^L_{k+1|k}; I_{L}] (P^L_{k+1|k} + \Sigma_{k+1})^{-1}$ 
\State $\tilde{z}_{k+1|k+1} = \tilde{z}_{k+1|k} + K_{k+1}(\mu_{k+1} - \mu_{k+1|k})$ 
\State$P_{k+1|k+1} = P_{k+1|k} - K_{k+1}(P^L_{k+1|k} + \Sigma_{k+1})K^\intercal_{k+1}$
\end{algorithmic}
\label{algorithm_KF}
\end{algorithm}
\vspace{-0.2cm}

\noindent{\bf Detection of new situation}. After the update phase, at each time instant $k$, the predicted value of $z_{k}^{l, p}$ related to latent state component $l$ and particle $p$ is compared with the actual updated value, outputting a measure of innovation defined as:
\begin{equation}\label{eq:abnormalityValue} 
y_k = \min_{p}\frac{\sum_{l = 1}^{L}\big|\mu_{k|k}^{l, p} - \mu_{k|k-1}^{l,p}\big|}{L}.
\end{equation}
The innovation values of training video sequences are used to set a threshold defined as:
\begin{equation}\label{eq:threshold} 
thresh = \bar{y}_{train} + 3std(y_{train}),
\end{equation}
where $\bar{y}_{train}$ and $std(y_{train})$ are the mean and standard deviation of innovations from the training data respectively. When applying algorithm \ref{algorithm_KF} on testing data, frames producing innovation values above the threshold in Eq.\eqref{eq:threshold} are considered as new situations. Moreover, to avoid spurious innovation peaks, a temporal window of 3 frames is used, such that new situations are recognized only if 3 consecutive frames are above $thresh$.
\vspace{-0.35cm}
\subsection{Continual Learning}\label{continual_learning}
\vspace{-0.1cm}
The calculation of innovations facilitates defining a continual learning process where frames belonging to new situations are detected and stored. These identified frames are employed to perform a new training process as described in section \ref{trainingPhase}, involving a new VAE ($\mathit{VAE}^{i: = i+1}$). $\mu_{train}$ and $\sigma_{train}^2$ are employed as bottleneck features related to the images of a new situation  $X_{train}^{i: = i+1}$.

During the new testing phase, the outputted bottleneck features and learned feature variables $V^j$ of the different VAEs $\mathit{VAE}^{j}$ where $j = \{1,\dots,i+1\}$, are used together in a single A-MJPF. The particles in the A-MJPF are then distributed among all the available clusters and consequently among the various VAEs, i.e., situations. 

Since the bottleneck features among VAEs capture a different meaning, in the MJPF, particles assigned to a particular VAE's cluster cannot be reassigned to other VAEs' clusters. Therefore, some particles do not jump between VAEs, but they always remain attached to a particular VAE. The innovation measurement (see Eq.~\ref{eq:abnormalityValue}) is again estimated in order to detect additional new situations. 




\begin{figure*}[t]
\begin{center}
\includegraphics[width=\linewidth]{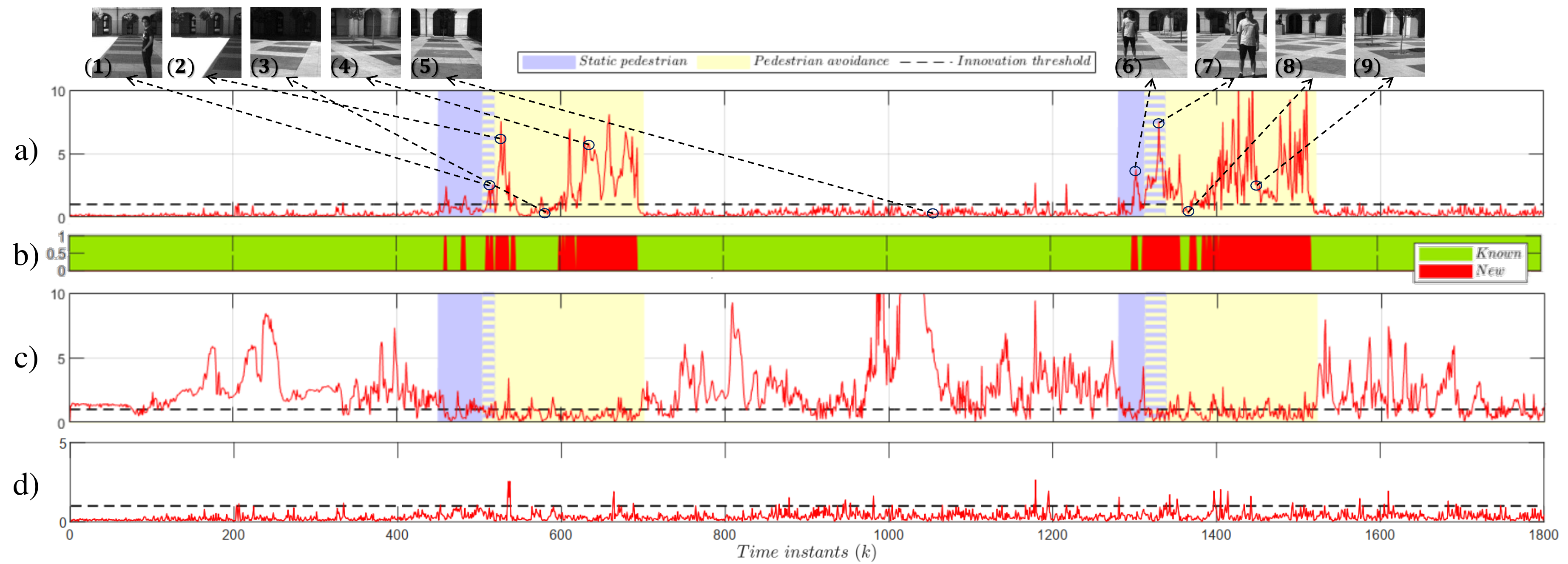}
\vspace{-0.65cm}
\caption{Testing phases on the pedestrian avoidance task: (a) Innovation signal using $\mathit{MODEL^{(PM)}}$. (b) Color-coded innovation. (c) Innovation signal using $\mathit{MODEL^{(PA)}}$. (d) Innovation signal using $\mathit{MODEL^{(PM, PA)}}$. }
\vspace{-0.75cm}
\label{fig:abn_OA}
\end{center}
\end{figure*}

\vspace{-0.25cm}
\section{Employed Dataset}\label{sec:dataset}
\vspace{-0.15cm}
A real vehicle called ``iCab'' \cite{Marin2016}, is used to collect video sequences from an onboard front camera. A human drives the iCab performing different tasks in a closed environment. 

This work aims at studying situations that have not been previously seen in a normal situation (Scenario I), which is used for learning purposes. Scenario II includes unseen maneuvers caused by the presence of pedestrians while the vehicle performs a previously seen task. The two scenarios considered in this work are:\\
\noindent{\bf Scenario I (perimeter monitoring)}. The vehicle follows a rectangular trajectory around a closed building. 

\begin{figure}[H]
  \centering
  \centerline{\includegraphics[width = \linewidth]{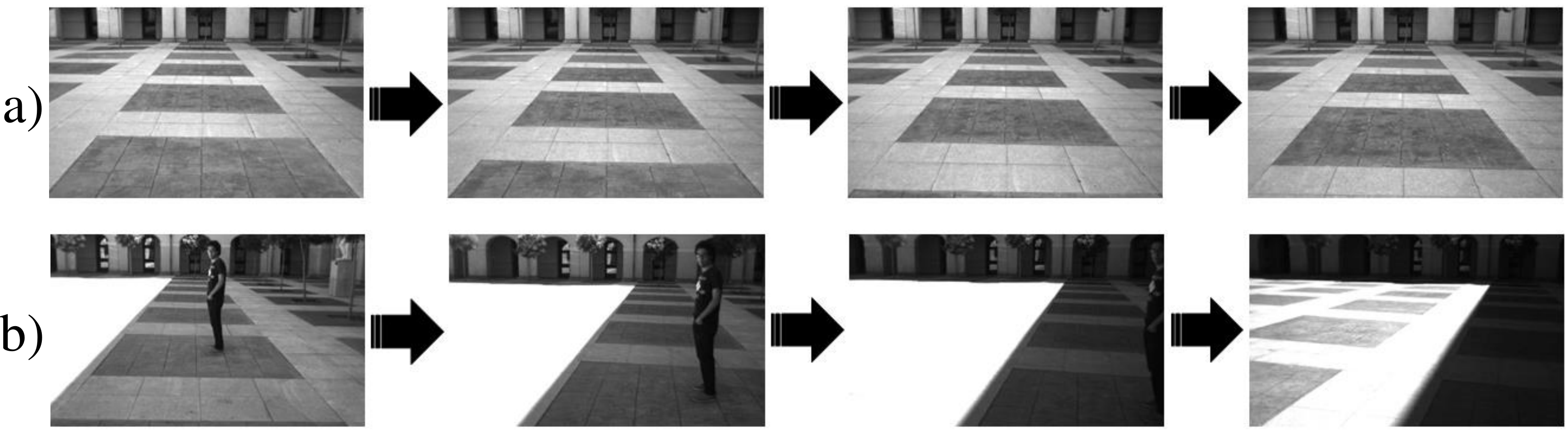}}
\caption{Video sequences from considered scenarios: a) parameter monitoring task and b) pedestrian avoidance maneuver.}
\vspace{-0.35cm}
\label{fig:scenarios}
\end{figure}

\noindent{\bf Scenario II (pedestrian avoidance maneuver)}. Two obstacles (stationary pedestrians) in different locations interfere with the perimeter monitoring task of Scenario I. The vehicle performs an avoidance maneuver and continues the perimeter monitoring. Fig.~\ref{fig:scenarios} shows a temporal evolution of video frames from both scenarios.
\vspace{-0.25cm}

\section{Experimental results}\label{sec:exp_results}
\vspace{-0.15cm}
\noindent{\bf First training phase: Perimeter Monitoring}. Initial models are trained based on the video frames from Scenario I. This data corresponds to $X^{i=1}_{train}$ and facilitates the obtainment of $\mathit{VAE}^{(1)}$ and the set of features in $V^{(1)}$, see section \ref{trainingPhase}. The threshold in Eq.\eqref{eq:threshold} is then calculated based on the behavior of the training data on the initial A-MJPF. We call $\mathit{MODEL^{(PM)}}$ the model using $\mathit{VAE}^{(1)}$ and the A-MJPF based on $V^{(1)}$.\\
\noindent{\bf Detection of new situation: Pedestrian Avoidance}. The icab faces a new situation: it encounters and avoids a static pedestrian. Fig. \ref{fig:abn_OA}a) shows the resulting innovation signal from $\mathit{MODEL^{(PM)}}$; blue zones refer to video frames containing the pedestrian and yellow regions encode the avoidance maneuvers. At each lap, the vehicle encounters two different static pedestrians, see images \textbf{(1)} and \textbf{(6)} in Fig. \ref{fig:abn_OA}a). They wear t-shirts of different colors (black and white), which make them ``camouflage'' with the environment in some particular configurations due to changeable illumination conditions. This factor influences the innovation values of frames \textbf{(1)} and \textbf{(6)}, with the second one generating a higher values.

Each maneuver of pedestrian avoidance generates two peaked zones, see frames\textbf{(2)} and \textbf{(4)} or \textbf{(7)} and \textbf{(9)}. Between such peaks, there is a zone with low innovation values, see \textbf{(3)} or \textbf{(8)}, due to the execution of similar behaviors already observed in the training set.

As described in section \ref{testingPhase}, the amplitude threshold obtained from the initial experience and a temporal window of 3 frames are used to detect the new situations. Fig. \ref{fig:abn_OA}b) displays the frames that were classified as known experiences (green) or new situations (red).

\vspace{-0.35cm}
\subsection{Learning of the new situation}
\vspace{-0.1cm}
The frames classified as new are used as $X^{i=2}_{train}$ for generating an additional VAE ($\mathit{VAE}^{(2)}$) and set of feature variables $V^{(2)}$ which can be used for generating a model that understands only the pedestrian avoidance maneuver; we call it $\mathit{MODEL^{(PA)}}$. Innovation measurements from $\mathit{MODEL^{(PA)}}$ are displayed in Fig. \ref{fig:abn_OA}c) where low innovations are obtained in zones related to the pedestrian presence and vehicle's avoidance maneuver. 
 
Innovation signals from $\mathit{MODEL^{(PM)}}$ and $\mathit{MODEL^{(PA)}}$ can be seen as complementary information that facilitates the incremental understanding of the proposed two tasks together, see how large innovations values in Fig. \ref{fig:abn_OA}a) correspond to low innovations in c). Accordingly, by employing available VAEs and variables: $\mathit{VAE}^{(1)}$, $\mathit{VAE}^{(2)}$, $V^{(1)}$, and $V^{(2)}$, we generate a single A-MPJF, called $\mathit{MODEL^{(PM,PA)}}$ that uses all previously learned concepts for prediction purposes, see section \ref{continual_learning}. Fig.\ref{fig:abn_OA}d) shows how innovations from $\mathit{MODEL^{(PM,PA)}}$ remain low through the entire scenario II, confirming the continual learning of new experiences avoiding the catastrophic forgetting of previously learned concepts. The percentage of false positive alarms of $\mathit{MODEL^{(PM,PA)}}$ is $7.14\%$.    
 
 
 \begin{figure}[t]
 \centering
 \includegraphics[width=\linewidth]{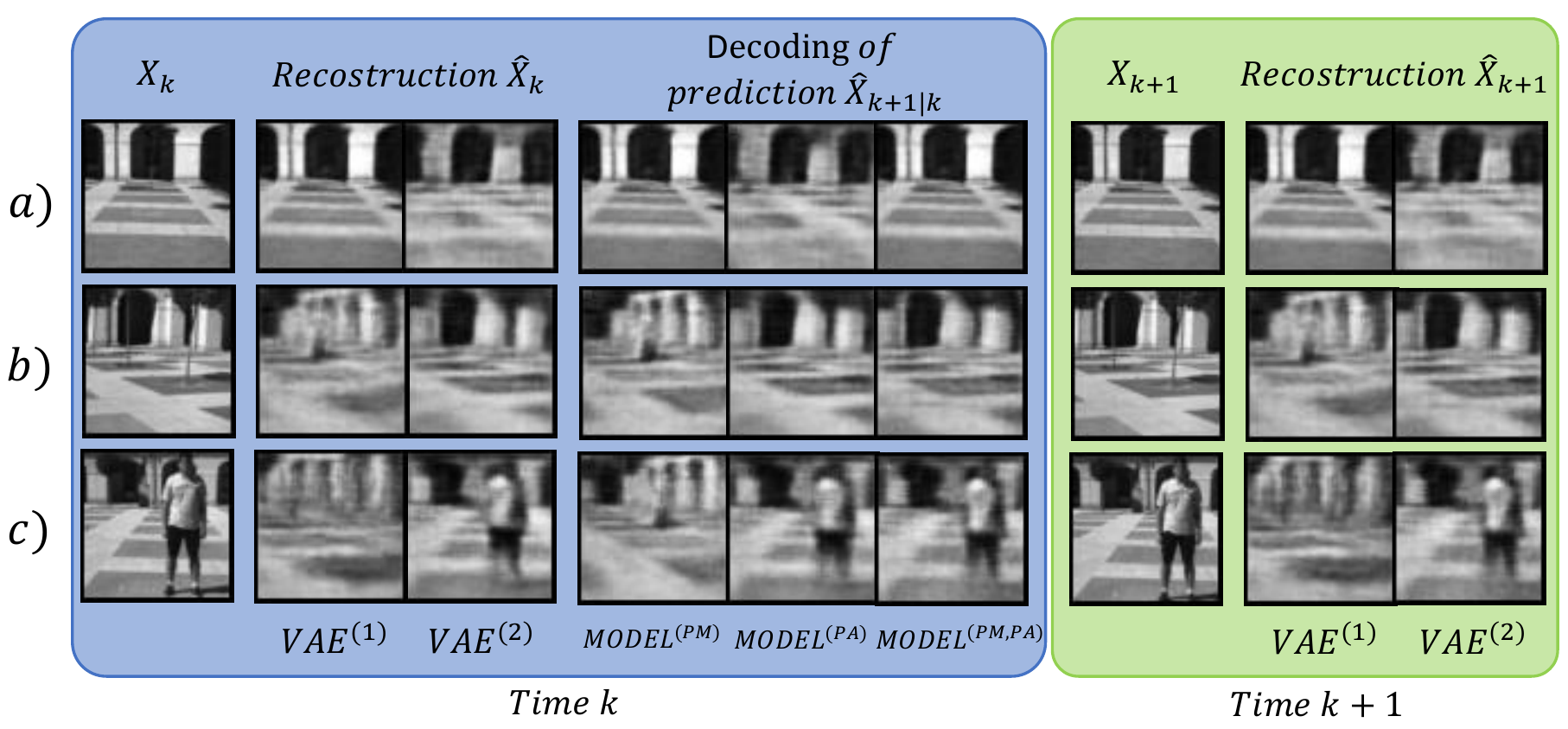}
 \vspace{-0.6cm}
 \caption{Examples of image-level behaviour in perimeter monitoring (a) and pedestrian avoidance (b-c) cases. } \vspace{-0.6cm}
 \label{fig:cases}
\end{figure}
Fig. \ref{fig:cases} displays three examples that explain the performance of our algorithm visually. From left to right the columns of the blue block correspond to: the image at time $k$; its reconstruction by using respectively $\mathit{VAE}^{(1)}$ and $\mathit{VAE}^{(2)}$; the decoded version of the predicted image at time $k+1$ given the image at time $k$ when adopting respectively $\mathit{MODEL^{(PM)}}$, $\mathit{MODEL^{(PA)}}$ and $\mathit{MODEL^{(PM, PA)}}$. Similarly, columns of the green block represent: the image at $k+1$; its reconstruction when using respectively $\mathit{VAE}^{(1)}$ and $\mathit{VAE}^{(2)}$. 
 
The row a) corresponds to the perimeter monitoring task, whereas rows b) and c)  are related to the pedestrian avoidance situation. $X_{k+1}$ and innovation values correspond to the ones shown in Fig. \ref{fig:abn_OA}a) with a blue circle, see frames \textbf{(5)}, \textbf{(7)}, \textbf{(9)}. Note how in a), the prediction of $\mathit{MODEL^{(PM)}}$ is accurate, while the one of $\mathit{MODEL^{(PA)}}$ is not, due to the wrong reconstruction of the image. In cases b) and c), the prediction of $\mathit{MODEL^{(PA)}}$ is accurate and the prediction of $MODEL^{(PM)}$ is not. This lousy performance of $\mathit{MODEL^{(PM)}}$ in case c) is again due to the image not being recognized, leading to inconsistencies while predicting. In the case of b), $\mathit{VAE}^{(1)}$ can generalize the observed image to a similar one that was in the training set. However, the prediction still produces high innovations because of a discrepancy between the expected motion and the observed dynamics: instead of moving left, the vehicle moves right to finalize the pedestrian avoidance maneuver. It can be visually observed how the prediction of $\mathit{MODEL^{(PM, PA)}}$ performs well in all three cases.
\vspace{-0.35cm}
\section{Conclusion and future work}
\vspace{-0.15cm}
\label{sec:conclusion}
The proposed work proposes a method that facilitates the continual learning of dynamical situations in video data. The proposed method is based on a probabilistic approach that uses latent spaces from VAEs to represent the state of video frames at each time instant. The dynamics of video sequences are captured by a set of NNs that encode different types of video motions in a given task. Future work includes the insertion of multimodal data into the A-MPJF, allowing the model to make inferences by fusing heterogeneous sensory data, e.g., video and positional information. Another possible path of the proposed work consists of improving the clustering process of latent space information, such that richer semantics can be obtained.


\bibliographystyle{IEEEbib}
\bibliography{refs}

\end{document}